\documentclass{article} 
\usepackage{iclr2027_conference,times}

\usepackage{amsmath,amsfonts,bm}

\def\eqref#1{equation~\ref{#1}}

\def\1{\bm{1}}

\DeclareMathAlphabet{\mathsfit}{\encodingdefault}{\sfdefault}{m}{sl}
\SetMathAlphabet{\mathsfit}{bold}{\encodingdefault}{\sfdefault}{bx}{n}

\usepackage{hyperref}
\usepackage{url}
\usepackage{algorithm}
\usepackage{algorithmic}
\usepackage{amsmath}
\usepackage{amssymb}
\usepackage{subcaption}
\usepackage{multirow}
\usepackage{booktabs}
\usepackage{xcolor}     
\usepackage{colortbl}
\usepackage{newfloat}
\usepackage{amssymb}
\usepackage{pifont}
\usepackage{listings}
\usepackage{graphicx}

\title{ Distilling Dense Trajectory Teachers for Large-Motion Video World Models}

\author{Shuofeng Sun\textsuperscript{1}, Linqing Zhao\textsuperscript{1}, Haibin Yan\textsuperscript{1}\thanks{Corresponding author}\\
Beijing University of Posts and Telecommunications\textsuperscript{1} \\
}

\iclrfinalcopy 
\begin{document}

\maketitle

\begin{abstract}
Video world models are expected to generate videos that accurately follow specified camera trajectories while maintaining
  consistent 3D structures. However, existing models are predominantly trained on densely sampled videos with small inter-frame
  camera motions. When applied to sparse, large-step trajectories, they often produce frozen frames, inaccurate camera motion, and
  severe geometric drift. We attribute this degradation to a mismatch between the gradual viewpoint transitions seen during
  training and the large viewpoint changes required at inference time. To address this problem, we propose Dense Trajectory Teacher
  Distillation (DTTD), a post-training framework that transfers a model’s generation capability from dense trajectories to large
  camera motions. Given a sparse trajectory, DTTD interpolates additional camera poses to construct a dense trajectory covering the
  same path. The frozen base model then generates reliable teacher videos along this familiar dense trajectory, and pose-aligned
  teacher observations are used to evaluate candidate videos generated under the original sparse trajectory. In this way, the model
  serves as its own generation teacher, requiring neither additional video data nor a stronger external video generator. We further
  combine dense-teacher supervision with geometry-aware rewards that assess cross-frame consistency and camera-trajectory
  following, and use the resulting preferences for offline preference optimization. To stably expand the model’s motion range, we
  introduce a cumulative curriculum that progressively incorporates larger motion steps while retaining simpler trajectories to
  reduce capability forgetting. Experiments demonstrate that DTTD effectively alleviates frozen-frame degradation, improves
  trajectory following and cross-frame geometric consistency under large-step camera motions, and preserves the visual quality of
  the base model.

\end{abstract}

\section{Introduction}

Recent video generation models~\cite{wan2025wan,hong2022cogvideo,yang2025cogvideox,sora} have achieved
  strong visual quality, temporal coherence, and long-horizon synthesis. As these models increasingly
  serve as world models for predicting future environments, camera-controllable variants generate videos
  from an initial image and a prescribed camera trajectory. Beyond photorealism, they must accurately
  follow the camera motion and preserve consistent scene geometry across frames. These capabilities are
  essential for reliable environment rollouts in embodied
intelligence~\cite{agarwal2025cosmos,team2025gigaworld,GR1}, autonomous driving~\cite{drivewm,wang2024drivedreamer,zhao2025drivedreamer2,gao2024vista}, and interactive scene exploration~\cite{team2026dreamx,matrixgame,lingbotworld}.

Recent studies have improved the 3D capabilities of video world models through post-training. Epipolar-DPO~\cite{epipolar-dpo}
  and VideoGPA~\cite{du2026videogpa} introduce geometric rewards based on epipolar constraints and geometry foundation models,
  while World-R1~\cite{world-r1} further incorporates vision-language feedback. These methods show that explicit geometric
  supervision can improve camera control and cross-frame consistency. However, they largely assume that test trajectories remain
  close to the training distribution and pay limited attention to the magnitude of camera motion.

\begin{figure}[!t]
    \centering
    \includegraphics[width=0.9\linewidth]{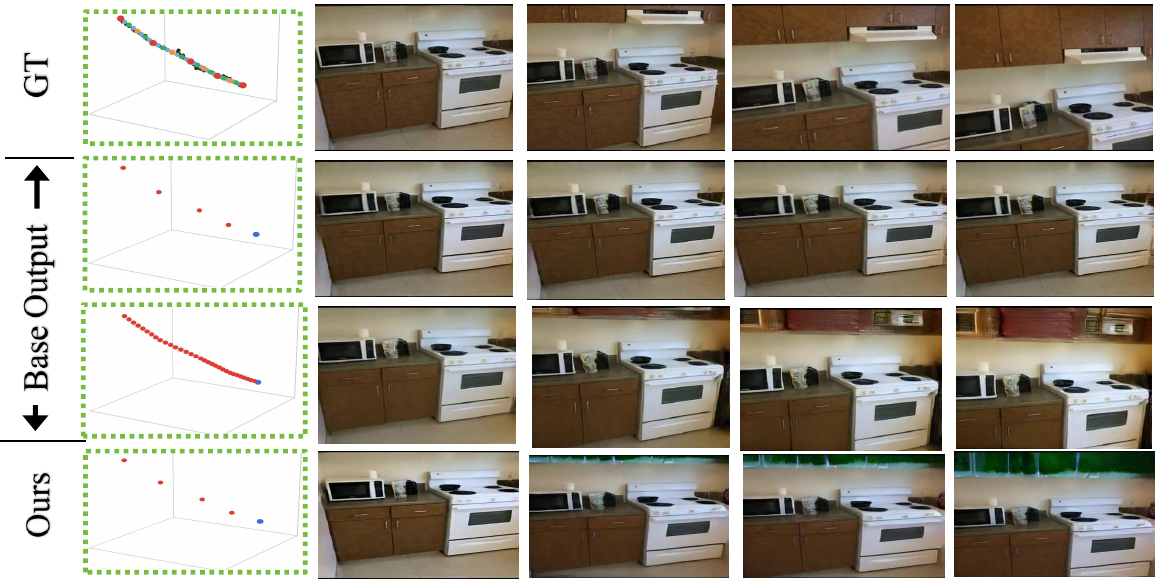}
    \caption{Generation video examples. When the camera trajectory is sparse and the motion magnitude is large, the model may produce static frames, geometric collapse, camera-following failure, and other issues. Furthermore, we can see that by densifying the trajectory and reducing the motion magnitude, the model's output quality can be substantially improved, which demonstrates the rationality of our dense trajectory evaluation. After alignment with our method, the model's geometric consistency, following capability, and visual quality are effectively improved when faced with large-motion trajectories.}
   
    \label{fig:teaser}
\end{figure}

  We find that the same model behaves very differently under dense and sparse sampling of the same camera path. It can generate
  coherent videos for dense, small-step trajectories, but often produces nearly static videos, fails to follow the specified
  motion, or exhibits severe geometric drift when adjacent poses are far apart, which is shown in Fig. \ref{fig:teaser}. This degradation is mainly caused by a trajectory-
  distribution mismatch: models are generally trained on gradual viewpoint changes, whereas large-step generation requires
  inferring substantial newly visible content and perspective variation in a single step. This raises a fundamental question: can
  the capability learned from dense trajectories be transferred to sparse, large-step camera motions?

To address this issue, we propose Dense Trajectory Teacher Distillation (DTTD), a post-training framework for video world models
  under large camera motions. Given a camera path, we construct a family of trajectories with different sampling rates, which
  follow the same path but exhibit progressively larger inter-frame motions. The densest trajectory, with the smallest motion
  steps, serves as the teacher trajectory. A frozen base model generates teacher videos along this trajectory, keeping each
  viewpoint transition within its familiar training distribution. For student trajectories of different motion difficulty, we
  extract teacher frames at their shared camera poses and use these pose-aligned observations to supervise the evaluation of
  student generations. We further introduce a cumulative curriculum that gradually progresses from small-step to large-step
  trajectories while retaining previously learned, simpler trajectories at each stage. This progressively expands the model’s
  motion range while mitigating capability forgetting. In this way, DTTD transfers the model’s existing dense-trajectory capability
  to camera motions of increasing magnitude without requiring additional video data or a stronger external video-generation model.

Although dense-trajectory teachers provide pose-aligned visual references, frame-level similarity alone
  cannot guarantee global 3D correctness or accurate camera control. We therefore use an external
  geometry foundation model to estimate scene geometry and camera motion. Each generated video is
  evaluated by relative-trajectory error after similarity-transform alignment with the input trajectory
  and by cross-frame reprojection consistency based on estimated depth and camera poses. These signals
  identify static or incorrectly moving generations, structural drift, inconsistent parallax, and
  deformation. Combined with dense-teacher supervision, they rank candidates, construct preference pairs,
  and guide offline preference optimization.
  

  Experimental results show that with the lightweight LoRA fine-tuning, our method can effectively improve the model's geometric consistency, camera-following capability, and visual quality for both small-step and large-step trajectories.

  \section{Related Works}
  \textbf{Video Generation Models.}  Most recent video generation models adopt Diffusion Transformers (DiT), which replace convolutional denoisers with Transformer
  blocks to model spatial-temporal tokens and scale effectively with data and model size~\cite{kong2024hunyuanvideo,sora,seedance2025seedance, agarwal2025cosmos,hong2022cogvideo}
  . To reduce the
  high computational cost of video diffusion, models such as CogVideoX~\cite{yang2025cogvideox} and Wan~\cite{wan2025wan} employ a causal 3D VAE
  that jointly compresses spatial and temporal dimensions before diffusion. This temporal compression relies on substantial
  redundancy between neighboring frames: consecutive frames are expected to contain similar appearance and geometry, allowing
  multiple frames to be represented by a shorter latent sequence. Consequently, these models are predominantly trained on densely
  sampled trajectories with small inter-frame camera transformations. When the trajectory is sparsely sampled and the camera moves
  by a large translation or rotation between adjacent frames, neighboring frames become less redundant and exhibit larger viewpoint
  and structural changes. Such inputs fall outside the effective training distribution of the temporally compressed model, often
  resulting in frozen motion, inaccurate trajectory following, and cross-frame geometric inconsistency.

\textbf{Video Diffusion Alignment.}
Post-training alignment of video diffusion models can further improve generation capability without training the model from scratch. DDPO~\cite{ddpo} models the diffusion sampling process as a sequential decision process and maximizes task rewards through policy optimization; Diffusion-DPO~\cite{diffusion-dpo} directly leverages human or synthetic preferences for DPO alignment without explicitly training a reward model; Flow-GRPO~\cite{flow-grpo} extends group-relative policy optimization to flow-matching generative models; and DanceGRPO~\cite{xue2025dancegrpo} applies relative-reward policy optimization to video diffusion models. These methods mainly provide generic feedback such as visual quality, motion quality, or semantic consistency, and are not specifically designed for 3D structural consistency and camera control. Accordingly, VideoGPA~\cite{du2026videogpa} employs geometry foundation models to evaluate cross-frame geometric consistency and camera motion, Epipolar-DPO~\cite{epipolar-dpo} constructs preference pairs based on epipolar constraints, and World-R1~\cite{world-r1} further incorporates vision-language model feedback. 


\section{Method}

\begin{figure}[!t]
    \centering
    \includegraphics[width=0.9\linewidth]{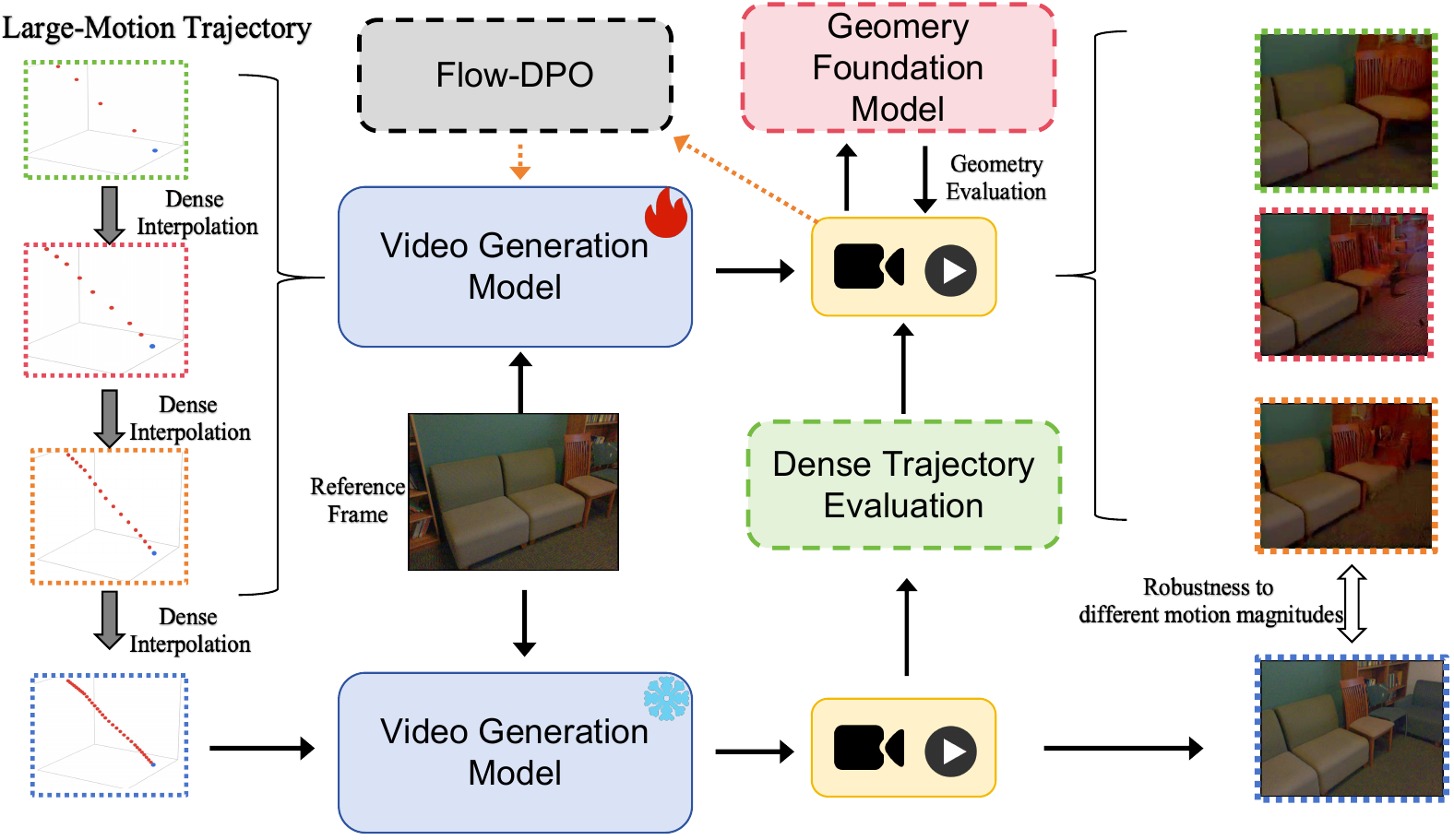}
    \caption{Pipeline of DTTD. We construct camera trajectories with different motion magnitudes through interpolation, and design dense trajectory evaluation and geometric evaluation to improve the model's performance when faced with large-motion trajectories.}
    \label{fig:model}
\end{figure}

We first introduce the task definition of camera-conditioned video generation and the basic objective of Flow-DPO in Sec. \ref{sec:1}. Then, Sec. \ref{sec:2} describes how to construct trajectories with different motion magnitudes along the same camera path, and how to use a frozen base model to generate dense-trajectory references for evaluating generation results under sparse trajectories. Sec. \ref{sec:3} further introduces an external geometry foundation model to provide global 3D evaluation from two aspects, cross-frame geometric consistency and camera trajectory following, so as to complement the pose-aligned dense teacher signal. Finally, Sec. \ref{sec:4} presents the construction of preference pairs and the cumulative curriculum training. The overall framework is shown in Fig. \ref{fig:model}.
\subsection{Preliminaries}
\label{sec:1}  
\textbf{Camera-Conditioned Video Generation.}
Let $I_0$ denote an initial image and $\mathcal{P}=\{P_t\}_{t=0}^{T-1}$ a camera trajectory containing $T$ poses, where
  $P_t\in SE(3)$ specifies the camera pose at frame $t$.  Then the camera-conditioned video model $\theta$, generates a video $X=\{I_t\}_{t=0}^{T-1}$

\textbf{Flow-DPO.}
For offline preference optimization, each training instance is represented as $\mathcal{P},X^{+},X^{-}$, where
  $X^{+}$ and $X^{-}$ are respectively a preferred and a dispreferred video generated under the same condition
  $\mathcal{P}$. Their clean latent representations are denoted by $\mathbf{z}^{+}_0=\mathcal{E}(X^{+})$ and $\mathbf{z}
  ^{-}_0=\mathcal{E}(X^{-})$.

  Let $\ell_\theta^{+}$ and $\ell_\theta^{-}$ denote the flow-matching errors of the trainable model on the preferred and
  dispreferred samples. Similarly, $\ell_{\mathrm{ref}}^{+}$ and $\ell_{\mathrm{ref}}^{-}$ denote the corresponding errors of a
  frozen reference model $\theta_{\mathrm{ref}}$. Then define $A_\theta^{\pm}=\ell_{\mathrm{ref}}^{\pm}-\ell_\theta^{\pm}$ as the
  improvement of the trainable model relative to the reference model. Flow-DPO encourages $A_\theta^{+}$ to be larger than
  $A_\theta^{-}$, such that the trainable model assigns a greater relative preference to $X^{+}$. The coefficient $\beta$
  controls the strength of preference optimization, and $\operatorname{sigmoid}(\cdot)$ denotes the sigmoid function. The overall loss can be defiend as follows:
  
  \begin{equation}
  \begin{aligned}
  \ell_{\theta}^{\pm}
  &=
  \left\|
  \mathbf{v}_{\theta}
  \left(
  \mathbf{z}_{\tau}^{\pm},
  \tau,
  \mathbf{c}
  \right)
  -
  \mathbf{v}^{*\pm}
  \right\|_2^2, \\
  A_{\theta}^{\pm}
  &=
  \ell_{\mathrm{ref}}^{\pm}
  -
  \ell_{\theta}^{\pm}, \\
  \mathcal{L}_{\mathrm{DPO}}(\theta)
  &=
  -\mathbb{E}_{(\mathbf{c},\mathbf{X}^{+},\mathbf{X}^{-})}
  \left[
  \log
  \operatorname{sigmoid}
  \left(
  \beta
  \left(
  A_{\theta}^{+}
  -
  A_{\theta}^{-}
  \right)
  \right)
  \right].
  \end{aligned}
  \label{eq:flow_dpo}
  \end{equation}

 Here, $c=(I_0,\mathcal{P})$ denotes the initial image and camera trajectory. $v_\theta$ and
  $v_{\mathrm{ref}}$ are the trainable and frozen reference velocity fields, $v^*$ is the target
  velocity, $\beta$ controls the preference strength, and $\sigma$ is the sigmoid function.
\subsection{Dense Trajectory Teachers.}
\label{sec:2}
\textbf{Multi-motion-scaled Trajectories Construction.}
  Given an original camera trajectory $\mathcal{P}=\{P_t\}{t=0}^{T-1}$, we construct a family of trajectories that follow the
  same camera path but exhibit different inter-frame motion magnitudes by varying the temporal sampling interval $s$. Specifically,
  we resample the discrete poses $P_t$ as $P_{st}$, where $0\leq st\leq T-1$. When $st$ is non-integral,
  $P_{st}$ is obtained by interpolating its two neighboring poses in the original trajectory. Since $s$ changes only the
  sampling locations along the path, all resulting trajectories share the same starting pose, ending pose, and underlying camera
  path.

  The original trajectory corresponds to $s=1$. When $s<1$, additional poses are inserted between neighboring original poses,
  producing a denser trajectory with smaller inter-frame motion. For example, $s=0.5$ doubles the sampling rate of the original
  trajectory. Conversely, $s>1$ yields a sparser trajectory with larger camera transformations between adjacent frames, resulting
  in a more difficult generation condition. We therefore define a trajectory with motion interval $s$ as:
  
  \begin{equation}
  \mathcal{P}^{(s)}
  =
  \left\{
  P_{st}
  \;\middle|\;
  t=0,\ldots,T_s-1
  \right\},
  \qquad
  T_s=\frac{T-1}{s}+1,
  \qquad
  0\leq st\leq T-1,
  \label{eq:motion_scaled_trajectory}
  \end{equation}

  To obtain the camera pose at a non-integral location $u=st$, we define $i=\lfloor u\rfloor$ and $\alpha=u-i$. We represent the
  $i$-th camera pose as $P_i=(\mathbf{R}_i,\mathbf{q}_i)$, where $\mathbf{R}_i\in SO(3)$ and $\mathbf{q}_i\in\mathbb{R}^3$
  denote its rotation and translation, respectively. The translation is linearly interpolated between $\mathbf{q}_i$ and $\mathbf{q}_
  {i+1}$, while the rotation is interpolated along the geodesic connecting $\mathbf{R}_i$ and $\mathbf{R}_{i+1}$ on $SO(3)$:

  \begin{equation}
      \mathbf{q}(u)
  =
  (1-\alpha)\mathbf{q}_i
  +
  \alpha\mathbf{q}_{i+1}
  \end{equation}

\begin{equation}
    \mathbf{R}(u)
  =
  \mathbf{R}_i
  \operatorname{Exp}
  \left(
  \alpha\,
  \operatorname{Log}
  \left(
  \mathbf{R}_i^{\top}\mathbf{R}_{i+1}
  \right)
  \right)
\end{equation}
 This construction ensures that all trajectories traverse the same continuous camera path, while $s$ controls only the motion
  magnitude between adjacent video frames. We use the trajectory with the smallest sampling interval $s_{teacher}$ as the teacher trajectory and
  trajectories with larger $s_{student}$ as student conditions of increasing difficulty.

\textbf{Dense-Trajectory Teacher Evaluation.}
Because adjacent poses along the teacher
  trajectory $s_{teacher}$ involve only small camera motions, this trajectory remains closer to the training distribution of the base model. We
  therefore use the frozen base model $\theta_{\mathrm{ref}}$ to generate $M$ teacher videos along this trajectory, denoted by
  $\{X^{teacher}_m\}_{m=1}^{M}$. For a student trajectory $\mathcal{P}^{(S_{student})}$ with larger inter-frame motions, we
  generate N student candidates $\{X^{student}_n\}_{n=1}^{N}$ using the same initial image and the corresponding sparse camera
  condition.

  Although the teacher and student trajectories have different sampling densities, they cover exactly the same camera path. Each
  pose in the student trajectory therefore has a corresponding pose in the teacher trajectory. The $t$-th student frame is located
  at $s_{student}t$ along the original path, and its corresponding teacher-frame index is $\pi(t)$.
  \begin{equation}
        \pi(t)=\frac{s_{student}}{s_{teacher}}t
  \end{equation}
This correspondence allows us to directly evaluate the scene observations produced by the model upon reaching the specified camera poses, without being affected by viewpoint bias caused by different trajectory sampling rates.

At each shared pose, we compare the student frame and the teacher frame from two aspects: geometric structure and visual content. First, we use a depth estimation model (VGGT~\cite{wang2025vggt}) to extract the depth maps of the two frames separately. Considering the scale ambiguity of monocular depth, we first perform scale alignment on the teacher depth and then compute the symmetric relative depth error. This depth distance is used to measure whether the scene layout and geometric structure described by the two frames are consistent.

  \begin{equation}
  \begin{aligned}
  \gamma_{t,n,m}^{*}
  &=
  \underset{\gamma>0}{\arg\min}\,
  \left\|
  \mathbf{D}_{t}^{n}
  -
  \gamma\hat{D}^{m}_{\pi(t)}
  \right\|_1, \\
  d_{\mathrm{depth}}^{(t,n,m)}
  &=
  \frac{1}{|\mathcal{Q}|}
  \sum_{\mathbf{q}\in\mathcal{Q}}
  \frac{
  \left|
  \mathbf{D}_{t}^{n}(\mathbf{q})
  -
  \gamma_{t,m}^{*}
  \hat{D}^{m}_{\pi(t)}(\mathbf{q})
  \right|
  }{
  \mathbf{D}_{t}^{n}(\mathbf{q})
  +
  \gamma_{t,m}^{*}
  \hat{D}^{m}_{\pi(t)}(\mathbf{q})
  +\epsilon
  },
  \end{aligned}
  \label{eq:teacher_depth_distance}
  \end{equation}
  where $\mathbf{D}_{t}^{n}$ is the depth of $t$-th frame in $n$-th student video.  $\hat{D}^{m}_{\pi(t)}$ is the depth of $\pi(t)$-th frame in $m$-th teacher video.

Second, we use the video model’s VAE encoder to extract single-frame latent features and compute their cosine distance. Compared
  with direct pixel differences, latent features are more tolerant of local color and texture variations while still capturing
  discrepancies in scene content and overall appearance.
  \begin{equation}
  d_{\mathrm{feat}}^{(t,n,m)}
  =
  1-
  \frac{
  \left\langle
  \mathbf{F}^{n}_t,
  \hat{F}^{m}_{\pi_(t)}
  \right\rangle
  }{
  \left\| \mathbf{F}^{n}_t\right\|_2
  \left\|\hat{F}^{m}_{\pi_(t)}\right\|_2
  },
  \label{eq:teacher_feature_distance}
  \end{equation}
  where $F$ and $\hat{F}$ mean the student and teacher single-frame features extracted by the video VAE encoder.

  Finally, we average both distances over all shared camera poses and the $M$ teacher videos, and combine them into the dense-trajectory
  teacher reward $R_{\mathrm{DT}}^n$ for $n$-th student video. A higher reward indicates that the observations generated by the student under large motion
  steps are closer to the reliable observations produced by the base model along the dense trajectory.
  \begin{equation}
      R_{\mathrm{DT}}^{n}= -\lambda_{\mathrm{depth}}\frac{1}{M|\Omega_s|}
  \sum_{m=1}^{M}
  \sum_{t\in\Omega_s}
  d_{\mathrm{depth}}^{(t,n,m)}-\lambda_{\mathrm{feat}} \frac{1}{M|\Omega_s|}
  \sum_{m=1}^{M}
  \sum_{t\in\Omega_s}
  d_{\mathrm{feat}}^{(t,n,m)}
  \end{equation}
  where $\Omega_s$ is the set of frame indices corresponding to the camera trajectory $\mathcal{P}^{(s)}$.

\textbf{Cumulative Motion Curriculum.} 
Although the dense-trajectory teacher provides a common evaluation criterion for student trajectories with different motion
  magnitudes, directly optimizing the model on large-step trajectories remains challenging. When the inter-frame camera motion
  substantially exceeds the model’s original capability, most candidate videos may exhibit frozen content, trajectory deviation, or
  structural collapse. Reliable preference pairs consequently become scarce, leading to unstable optimization. Moreover, training
  exclusively on the largest motions may cause the model to forget its existing capability on simpler trajectories.

  We therefore use a cumulative motion curriculum. Let $s_1\leq\cdots\leq s_K$ denote increasingly large
  sampling intervals. At stage (k), we train on trajectories $\{s_1,\ldots,s_k\}$, mixing them with
  predefined weights. This progressively expands the motion range while retaining simpler trajectories
  for rehearsal. In our experiments, ($s_{teacher}=0.125)$. The three stages use (s=0.25), ({0.25,0.5}), and
  ({0.25,0.5,1}), respectively.




\subsection{Geometry Foundation Model Evaluation}
\label{sec:3}
 Dense-trajectory teachers evaluate whether the student produces plausible scene observations at shared camera poses, but
  frame-level matching alone cannot fully guarantee global 3D consistency or accurate trajectory following. We therefore introduce
  a frozen geometry foundation model to estimate per-frame depth and camera poses from each generated video and derive two
  complementary rewards.

\textbf{Cross-frame Geometric Consistency.}
 Using the estimated depths and camera poses, pixels from one frame are back-projected into 3D and reprojected into other frames.
  The discrepancy between the reprojected observation and the target frame measures whether the recovered scene remains consistent
  across viewpoints, exposing geometric drift, incorrect parallax, and structural deformation. We define the negative reprojection
  error as the geometric-consistency reward $R_{\mathrm{geo}}$.

\textbf{Trajectory Following Reward.} 
 A geometrically consistent video does not necessarily follow the conditioning trajectory, and a nearly static video may even
  yield a small cross-frame error. We therefore compare the estimated camera trajectory $\widehat{\mathcal{P}}$ with the
  conditioning trajectory $\mathcal{T}$. Because monocular geometry estimation is ambiguous up to a global rotation, translation,
  and scale, the two trajectories are first aligned using a $\operatorname{Sim}(3)$ transformation. Camera-center and rotational
  geodesic errors are then combined into the trajectory-following reward $R_{\mathrm{pose}}$, which identifies static generation
  and camera motions with incorrect directions or magnitudes.

  These signals respectively evaluate the internal 3D consistency of the video and the correctness of its motion relative to the
  input condition, complementing dense-trajectory teacher supervision.

  \subsection{Training}
 For each initial image and student trajectory condition $\mathcal{P}^{(s_k)}$, we generate $N$
  candidates and evaluate them using dense-teacher, geometric-consistency, and trajectory-following
  rewards. After within-group normalization, we combine these rewards to rank the candidates. The highest and lowest scoring videos form a preference pair, retained only when their reward gap exceeds $\delta$.

  We adopt a cumulative curriculum: stage $k$ uses preference data from $\{s_1,\ldots,s_k\}$, progressively
  expanding the motion range while retaining simpler trajectories to mitigate forgetting. The model is
  optimized offline with Flow-DPO by updating only its LoRA parameters.


\label{sec:4}
\section{Experiments}
\subsection{Experiment Setup}
\textbf{Dataset.} We construct preference pairs on the ScanNet~\cite{dai2017scannet} training set, with a total of 10K pairs, and then evaluate the methods on the ScanNet test set. Furthermore, we use the trajectories provided by ScanNet for training, but instead of directly using the original trajectories as dense trajectory supervision, we downsample the original trajectories to obtain sparse, large-motion trajectories and then construct dense trajectories via interpolation. This provides a more generalizable post-training setting that does not require real dense trajectories.

\textbf{Implementation Details.} We adopt Wan2.1-Fun-V1.1-1.3B-Control-Camera as the camera-conditioned video generation base model. For each student condition, we generate $N$=12 candidate videos, and sample 4 teacher videos for each dense trajectory to reduce generation randomness. Video generation uses the UniPC solver, 40 denoising steps, a guidance scale of 5.0, and a flow shift of 5.0. Training adopts a three-stage cumulative curriculum. The first stage trains for 3,000 steps using only trajectories with $s=0.25$
; the second stage mixes $s\in \{0.25,0.5\}$ with equal probability and continues training for 3,000 steps; the third stage uses $s\in \{0.25,0.5,1\}$ with sampling weights of {
0.3, 0.3, 0.4
} and trains for 4,000 steps. We adopt Flow-DPO with $\beta=500$ and update only the LoRA modules with rank 64 and scaling factor 128. The optimizer is AdamW with a learning rate of 5e-6
 , a warmup of the first 500 steps followed by cosine decay, and a maximum gradient norm of 1.0.
\begin{table*}[t]
  \centering
  \caption{
  Visual-quality comparison on the ScanNet test set.
  s = 0.125 denotes the densest reference trajectory.
  Relative to this trajectory, s = 0.5 and s = 1 correspond to 4x and 8x larger sampling intervals,
  respectively.
  The best and second-best results are highlighted in
  \textbf{bold} and \underline{underline}.
  }
  \label{tab:vbench_comparison}
  \setlength{\tabcolsep}{5.5pt}
  \renewcommand{\arraystretch}{1.08}
  \resizebox{\textwidth}{!}{
  \begin{tabular}{clcccccc}
  \toprule
  Motion
  & Method
  & Subject Cons. \(\uparrow\)
  & Background Cons. \(\uparrow\)
  & Motion Smooth. \(\uparrow\)
  & Dynamic Degree \(\uparrow\)
  & Aesthetic Quality \(\uparrow\)
  & Imaging Quality \(\uparrow\) \\
  \midrule

  \multirow{4}{*}{\textbf{s=0.125}}
  & Base
  & 0.870108 & 0.928226 & 0.989094
  & \textbf{0.934535} & 0.339647 & 0.439500 \\
  & Epipolar-DPO
  & 0.885960 & 0.935076 & \textbf{0.991633}
  & 0.903226 & 0.345566 & 0.378853 \\
  & VideoGPA
  & \underline{0.889117} & \underline{0.936558} & 0.989903
  & 0.920304 & \underline{0.349204} & \textbf{0.455657} \\
  & \textbf{DTTD (Ours)}
  & \textbf{0.903036} & \textbf{0.942309} & \underline{0.991436}
  & \underline{0.924099} & \textbf{0.354848} & \underline{0.449773} \\
  \midrule

  \multirow{4}{*}{\textbf{s=0.5}}
  & Base
  & 0.815244 & 0.907723 & 0.975771
  & \textbf{0.988615} & 0.330851 & 0.388325 \\
  & Epipolar-DPO
  & 0.844630 & 0.919677 & \underline{0.980837}
  & 0.972486 & 0.336900 & 0.353353 \\
  & VideoGPA
  & \underline{0.849110} & \underline{0.920342} & 0.978041
  & \underline{0.984820} & \underline{0.340052} & \underline{0.409274} \\
  & \textbf{DTTD (Ours)}
  & \textbf{0.888350} & \textbf{0.934290} & \textbf{0.982276}
  & 0.976281 & \textbf{0.349278} & \textbf{0.413674} \\
  \midrule

  \multirow{4}{*}{\textbf{s=1}}
  & Base
  & 0.802263 & 0.904033 & 0.970840
  & \textbf{0.988615} & 0.334161 & 0.382421 \\
  & Epipolar-DPO
  & 0.833362 & \underline{0.917281} & \underline{0.976459}
  & 0.983871 & 0.339950 & 0.351915 \\
  & VideoGPA
  & \underline{0.836967} & 0.916863 & 0.973468
  & \underline{0.984201} & \underline{0.341548} & \underline{0.401181} \\
  & \textbf{DTTD (Ours)}
  & \textbf{0.885077} & \textbf{0.933819} & \textbf{0.979471}
  & 0.982922 & \textbf{0.351901} & \textbf{0.407055} \\
  \bottomrule
  \end{tabular}
  }
  \end{table*}

 \begin{table*}[!t]
  \centering
  \caption{
  Camera-following performance on the ScanNet test set.
  The best and second-best results in each setting are highlighted
  in \textbf{bold} and \underline{underline}.
  }
  \label{tab:camera_following}
  \setlength{\tabcolsep}{7pt}
  \renewcommand{\arraystretch}{1.08}
  \resizebox{\textwidth}{!}{
  \begin{tabular}{clccccc}
  \toprule
  Motion
  & Method
  & CamMC \(\downarrow\)
  & CamMC Rot. \(\downarrow\)
  & CamMC Trans. \(\downarrow\)
  & ATE. Trans. \(\downarrow\)
  & ATE. Rot. \(\downarrow\) \\
  \midrule

  \multirow{4}{*}{$s=0.125$}
  & Base
  & 0.518166 & 0.094355 & 0.488861 & 0.158650 & 1.730458 \\
  & Epipolar-DPO~\cite{epipolar-dpo}
  & 0.499797 & \underline{0.085560} & 0.474069
  & \textbf{0.146549} & 1.667481 \\
  & VideoGPA~\cite{du2026videogpa}
  & \underline{0.492305} & 0.085704 & \underline{0.466349}
  & \underline{0.149542} & \underline{1.651467} \\
  & \textbf{DTTD (Ours)}
  & \textbf{0.488283} & \textbf{0.076850} & \textbf{0.452347}
  & 0.158356 & \textbf{1.609799} \\
  \midrule

  \multirow{4}{*}{$s=0.5$}
  & Base
  & 0.660112 & 0.152059 & 0.601717 & 0.181642  & 2.029051 \\
  & Epipolar-DPO~\cite{epipolar-dpo}
  & 0.658929 & 0.140667 & 0.609915
  & 0.172629 & 2.020727 \\
  & VideoGPA~\cite{du2026videogpa}
  & \underline{0.640154} & \underline{0.139283} & \underline{0.590217}
  & \underline{0.171267} & \underline{1.973013} \\
  & \textbf{DTTD (Ours)}
  & \textbf{0.615385} & \textbf{0.120957} & \textbf{0.577336}
  & \textbf{0.167662} & \textbf{1.949126} \\
  \midrule

  \multirow{4}{*}{$s=1$}
  & Base
  & 0.712934 & 0.171629 & 0.646084
  & 0.178227 & 2.109135 \\
  & Epipolar-DPO~\cite{epipolar-dpo}
  & 0.700300 & 0.157462 & 0.644265 & 0.179254 & 2.107812 \\
  & VideoGPA~\cite{du2026videogpa}
  & \underline{0.685212} & \underline{0.156230} & \underline{0.628880}
  & \underline{0.175211} & \underline{2.074575} \\
  & \textbf{DTTD (Ours)}
  & \textbf{0.659684} & \textbf{0.141219} & \textbf{0.612444}
  & \textbf{0.166054} & \textbf{2.046553} \\
  \bottomrule
  \end{tabular}
  }
  \end{table*}

\subsection{Main Results}
In this section, we mainly report the visual quality, 3D consistency, and camera-following capability of different methods on ScanNet test under trajectories with different motion magnitudes. We mainly compare the Base Model~\cite{wan2025wan}, VideoGPA~\cite{du2026videogpa}, epipolar-dpo~\cite{epipolar-dpo}, and DTTD. It should be emphasized that VideoGPA, epipolar-dpo, and DTTD all apply preference optimization to the model using the same data and the same number of training steps, so as to ensure a fair comparison.

\textbf{Visual Quality.} 
In Tab.~\ref{tab:vbench_comparison}, we compare the visual quality of different methods under varying
  motion magnitudes using VBench~\cite{huang2024vbench}. DTTD achieves the best overall performance, with
  particularly clear gains under sparse trajectories ($s=0.5$ and $s=1$) in subject consistency,
  background consistency, aesthetic quality, and imaging quality. As the sampling interval increases from
  $s=0.125$ to $s=1$, the subject consistency of the base model drops from 0.8701 to 0.8023, while DTTD
  decreases only from 0.9030 to 0.8851, demonstrating greater robustness to trajectory sparsification.
  Although DTTD has a slightly lower Dynamic Degree than some baselines, this metric measures motion
  magnitude rather than trajectory accuracy. Combined with the camera-following results in
  Tab.~\ref{tab:camera_following}, these results show that DTTD produces more accurate, rather than
  merely stronger, motion while preserving visual quality.

\begin{table*}[!t]
  \centering

  \begin{minipage}[t]{0.48\textwidth}
  \vspace{0pt}
  \centering
  \captionof{table}{
  3D consistency comparison on the ScanNet test set.
  Metrics are computed on 10 uniformly sampled frames using VGGT,
  with SIFT used for epipolar matching.
  The best and second-best results are highlighted in
  \textbf{bold} and \underline{underline}.
  }
  \label{tab:3d_consistency}

  \setlength{\tabcolsep}{5pt}
  \renewcommand{\arraystretch}{1.08}
  \resizebox{\linewidth}{!}{
  \begin{tabular}{clccc}
  \toprule
  Motion & Method
  & 3DCS \(\downarrow\)
  & MVCS \(\uparrow\)
  & Epipolar \(\downarrow\) \\
  \midrule

  \multirow{4}{*}{$s=0.125$}
  & Base          & 0.700308 & 0.993523 & 4.185630 \\
  & Epipolar-DPO  & 0.660622 & \underline{0.997204} & \textbf{2.014262} \\
  & VideoGPA      & \underline{0.647028} & 0.996177 & 3.647692 \\
  & \textbf{DTTD (Ours)}
                  & \textbf{0.586834} & \textbf{0.997290} & \underline{2.758026} \\
  \midrule

  \multirow{4}{*}{$s=0.5$}
  & Base          & 0.696200 & 0.994591 & 4.113428 \\
  & Epipolar-DPO  & 0.657286 & \underline{0.996707} & \textbf{2.410791} \\
  & VideoGPA      & \underline{0.655636} & 0.995543 & 3.759172 \\
  & \textbf{DTTD (Ours)}
                  & \textbf{0.586626} & \textbf{0.997061} & \underline{3.513251} \\
  \midrule

  \multirow{4}{*}{$s=1$}
  & Base          & 0.691349 & 0.994027 & 3.989733 \\
  & Epipolar-DPO  & 0.654055 & \underline{0.996206} & \textbf{2.230878} \\
  & VideoGPA      & \underline{0.653449} & 0.995061 & 3.754776 \\
  & \textbf{DTTD (Ours)}
                  & \textbf{0.582469} & \textbf{0.997387} & \underline{3.201833} \\
  \bottomrule
  \end{tabular}
  }
  \end{minipage}
  \hfill
  \begin{minipage}[t]{0.48\textwidth}
  \vspace{0pt}
  \centering
  \includegraphics[width=\linewidth]{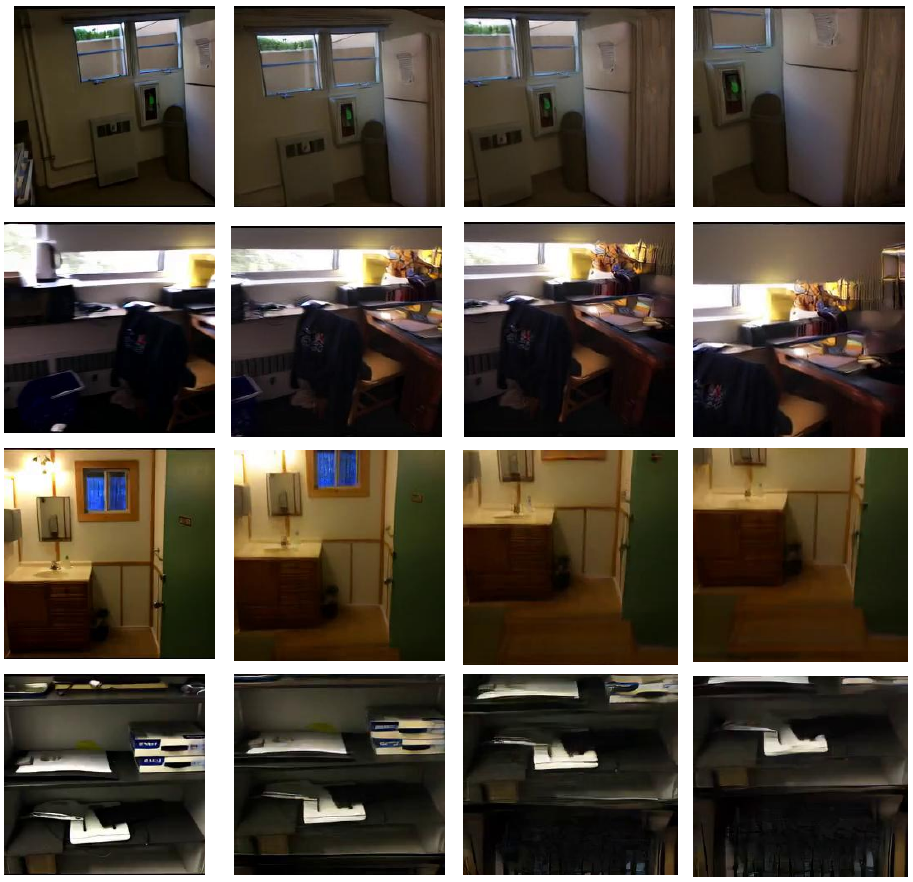}
  \captionof{figure}{Visualization of generation videos.}
  \label{fig:vis_gen}
  \end{minipage}

  \end{table*}

\textbf{Traject Following.} In Tab. \ref{tab:camera_following}, we report the trajectory following metrics of different methods under different motion magnitudes. We first report the CamMC, CamMc Rot and CamMc Trans from ~\cite{zheng2024cami2v,he2025cameractrl}. We also report the rotation error and translation error of Absolute Trajectory Error (ATE). The former is more stringent in measuring following capability, while the latter is more robust to monocular scale ambiguity. We can see that DTTD achieves the best CamMC, CamMC Rot, CamMC Trans, and ATE. Rot across all motion magnitudes, and its advantages are more pronounced at the larger motion, indicating that it can handle large inter-frame camera motion more robustly.

\textbf{3D Consistency.}In Tab. \ref{tab:3d_consistency}, we report the trajectory following metrics of different methods under different motion magnitudes.  We use three complementary metrics for 3D consistency. 3DCS measures pixel and perceptual errors
  between reprojected and target frames, where lower is better. MVCS evaluates depth-reprojection
  consistency across adjacent views, where higher is better. Epipolar computes the Sampson distance of
  cross-frame SIFT correspondences, where lower indicates better compliance with epipolar geometry.
  DTTD achieves the best 3DCS and MVCS across all motion magnitudes. Compared with the base model, it
  reduces 3DCS by approximately (16.2\%), (15.7\%), and (15.7\%) on $s=0.125$, $s=0.5$, and $s=1$, respectively,
  showing that its geometric improvements remain stable under sparse trajectories.
  Epipolar-DPO performs best on the Epipolar metric, consistent with its direct optimization of epipolar
  constraints. Although DTTD does not explicitly optimize this metric, it ranks second under all three
  motion settings and reduces the error over the base model by approximately (34.1\%), (14.6\%), and
  (19.7\%).

\begin{figure}
    \centering
    \includegraphics[width=0.9\linewidth]{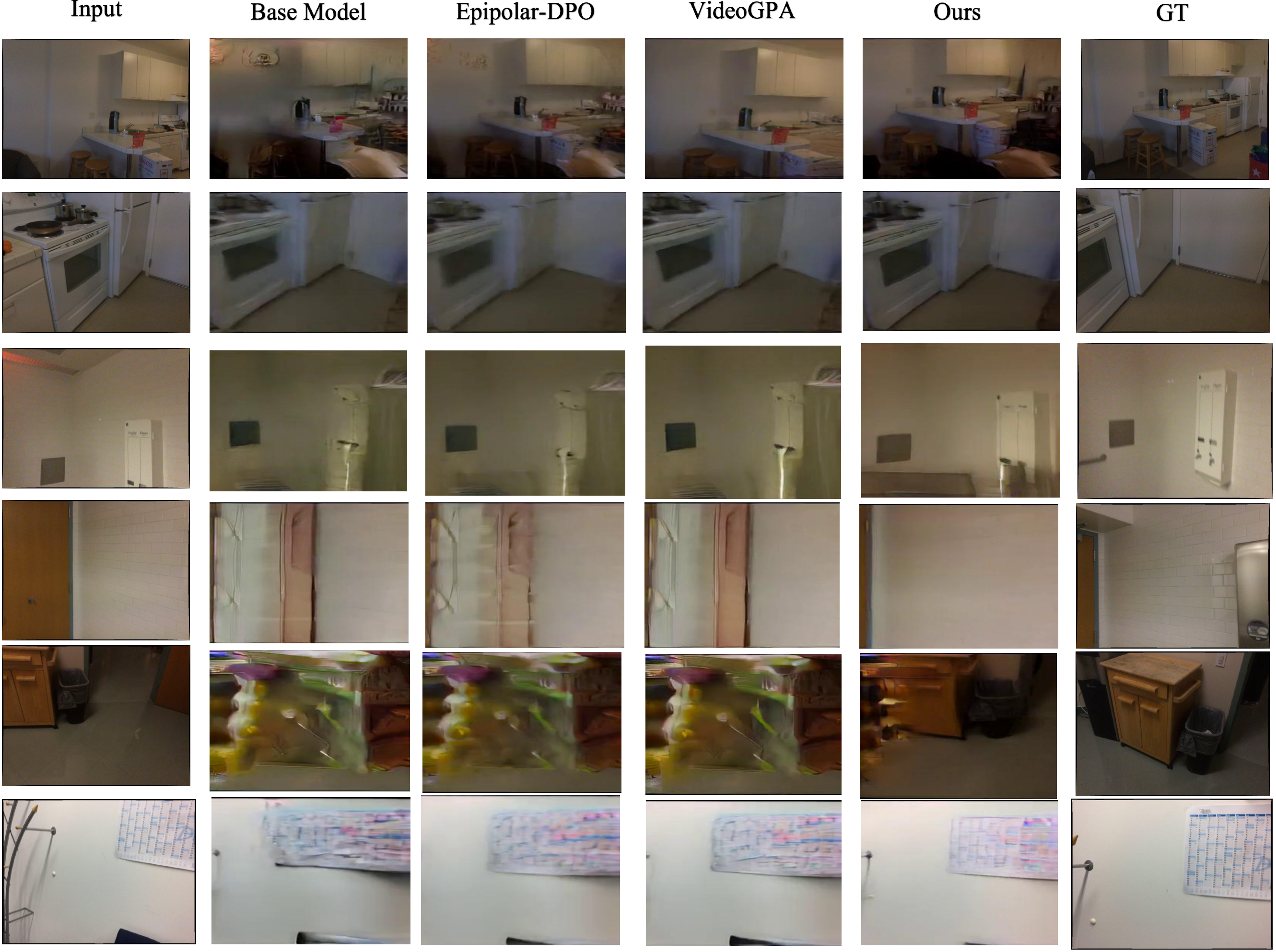}
    \caption{Qualitative comparison of Different Methods.}
    \label{fig:vis_compare}
\end{figure}

\textbf{Visualizations.} In Fig. \ref{fig:vis_gen}, we visualize the videos generated by our method. The first two rows show the generation results under the setting 
s=0.125, and the last two rows show the generation results under 
s=1. It can be seen that DTTD can effectively improve the model's generation quality under different motion magnitudes.  Furthermore, in Fig.\ref{fig:vis_compare}, we visualize the generation results of different methods when faced with s=0.125 (first three rows) and s=1 (last three rows).

 \begin{table*}[!t]
  \centering
  \caption{
  Cumulative ablation study under different trajectory sampling intervals.
  Each row adds one component to the preceding configuration. IQ means the Imaging Quality. 
  }
  \label{tab:cumulative_ablation}
  \setlength{\tabcolsep}{4.2pt}
  \renewcommand{\arraystretch}{1.1}
  \resizebox{\textwidth}{!}{
  \begin{tabular}{l|ccc|ccc|ccc}
  \toprule
  \multirow{2}{*}{Method}
  & \multicolumn{3}{c|}{s=0.125}
  & \multicolumn{3}{c|}{s=0.5}
  & \multicolumn{3}{c}{s=1} \\
  \cmidrule(lr){2-4}
  \cmidrule(lr){5-7}
  \cmidrule(lr){8-10}
  & 3DCS \(\downarrow\)
  & CamMC \(\downarrow\)
  & IQ \(\uparrow\)
  & 3DCS \(\downarrow\)
  & CamMC \(\downarrow\)
  & IQ \(\uparrow\)
  & 3DCS \(\downarrow\)
  & CamMC \(\downarrow\)
  & IQ \(\uparrow\) \\
  \midrule

  Base
  & 0.7003 & 0.5182 & \underline{0.4395}
  & 0.6962 & 0.6601 & 0.3883
  & 0.6913 & 0.7129 & 0.3824 \\

  + Dense-Trajectory Evaluation
  & 0.6254 & 0.5018 & 0.4352
  & 0.6368 & 0.6372 & 0.3995
  & 0.6342 & 0.6861 & 0.3936 \\

  + Geometry Evaluation
  & \underline{0.5967} & \underline{0.4919} & 0.4376
  & \underline{0.5979} & \underline{0.6216} & \underline{0.4078}
  & \underline{0.5948} & \underline{0.6675} & \underline{0.4018} \\

  + Cumulative Curriculum
  & \textbf{0.5868} & \textbf{0.4883} & \textbf{0.4398}
  & \textbf{0.5866} & \textbf{0.6154} & \textbf{0.4137}
  & \textbf{0.5825} & \textbf{0.6597} & \textbf{0.4071} \\
  \bottomrule
  \end{tabular}
  }
  \end{table*}

\subsection{Ablation Studies}
\textbf{Different Method Components.} In Table \ref{tab:cumulative_ablation}, we test the effect of different components of our method. Dense-trajectory evaluation substantially improves 3DCS and CamMC across all motion intervals,
  confirming that the model’s dense-trajectory capability can supervise large-step generation. Geometry
  evaluation provides further gains through reprojection and trajectory feedback. The cumulative
  curriculum achieves the best overall performance, particularly improving imaging quality on larger motion, showing that progressive motion expansion enhances large-motion generation while mitigating
  forgetting.


\section{Conclusion}
We presented DTTD, a post-training framework that transfers a video world model's dense-trajectory
  capability to sparse, large-step camera motions without additional video data or an external generation
  teacher. Combined with geometry-aware evaluation and cumulative curriculum learning, DTTD improves
  camera following and cross-frame 3D consistency while maintaining visual quality.



\bibliography{iclr2027_conference}
\bibliographystyle{iclr2027_conference}

\newpage
\appendix
\section{Appendix}
\begin{table}[!t]
    \centering
    \caption{
    Ablation of the dense-trajectory teacher reward.
    The depth-based term uses scale-aligned depth consistency, while the
    feature-based term measures the cosine distance between VAE latent features.
    Lower is better for 3DCS and CamMC, while higher is better for IQ.
    }
    \label{tab:dense_teacher_ablation}
    \setlength{\tabcolsep}{5pt}
    \renewcommand{\arraystretch}{1.05}
    \resizebox{\linewidth}{!}{
    \begin{tabular}{lccc|ccc|ccc}
      \toprule
      & \multicolumn{3}{c|}{$s=0.125$}
      & \multicolumn{3}{c|}{$s=0.5$}
      & \multicolumn{3}{c}{$s=1$} \\
      Method
      & 3DCS $\downarrow$ & CamMC $\downarrow$ & IQ $\uparrow$
      & 3DCS $\downarrow$ & CamMC $\downarrow$ & IQ $\uparrow$
      & 3DCS $\downarrow$ & CamMC $\downarrow$ & IQ $\uparrow$ \\
      \midrule
      Base
      & 0.7003 & 0.5182 & 0.4395
      & 0.6962 & 0.6601 & 0.3883
      & 0.6913 & 0.7129 & 0.3824 \\

      + Feature only
      & 0.6578 & 0.5074 & 0.4371
      & 0.6549 & 0.6468 & 0.3972
      & 0.6521 & 0.6964 & 0.3915 \\

      + Depth only
      & 0.6397 & 0.5031 & 0.4368
      & 0.6425 & 0.6349 & 0.4008
      & 0.6381 & 0.6817 & 0.3952 \\

      + Depth + Feature
      & \textbf{0.6254} & \textbf{0.5018} & \textbf{0.4352}
      & \textbf{0.6368} & \textbf{0.6372} & \textbf{0.3995}
      & \textbf{0.6342} & \textbf{0.6861} & \textbf{0.3936} \\
      \bottomrule
    \end{tabular}
    }
  \end{table}

  \begin{table}[t]
    \centering
    \caption{
    Ablation of the geometry-aware rewards.
    Reprojection evaluates cross-frame geometric consistency, while relative
    trajectory evaluates camera-motion following after similarity-transform
    alignment. Lower is better for 3DCS and CamMC, while higher is better for IQ.
    }
    \label{tab:geometry_reward_ablation}
    \setlength{\tabcolsep}{5pt}
    \renewcommand{\arraystretch}{1.05}
    \resizebox{\linewidth}{!}{
    \begin{tabular}{lccc|ccc|ccc}
      \toprule
      & \multicolumn{3}{c|}{$s=0.125$}
      & \multicolumn{3}{c|}{$s=0.5$}
      & \multicolumn{3}{c}{$s=1$} \\
      Method
      & 3DCS $\downarrow$ & CamMC $\downarrow$ & IQ $\uparrow$
      & 3DCS $\downarrow$ & CamMC $\downarrow$ & IQ $\uparrow$
      & 3DCS $\downarrow$ & CamMC $\downarrow$ & IQ $\uparrow$ \\
      \midrule
      Dense teacher only
      & 0.6254 & 0.5018 & 0.4352
      & 0.6368 & 0.6372 & 0.3995
      & 0.6342 & 0.6861 & 0.3936 \\

      + Reprojection reward
      & 0.6048 & 0.5009 & 0.4363
      & 0.6087 & 0.6318 & 0.4054
      & 0.6041 & 0.6769 & 0.3999 \\

      + Relative trajectory reward
      & 0.6172 & 0.4897 & 0.4360
      & 0.6205 & 0.6168 & 0.4059
      & 0.6176 & 0.6614 & 0.4007 \\

      + Both geometry rewards
      & \textbf{0.5967} & \textbf{0.4919} & \textbf{0.4376}
      & \textbf{0.5979} & \textbf{0.6216} & \textbf{0.4078}
      & \textbf{0.5948} & \textbf{0.6675} & \textbf{0.4018} \\
      \bottomrule
    \end{tabular}
    }
  \end{table}

\begin{table}[!t]
    \centering
    \caption{
    Ablation of different motion curriculum strategies.
    All variants use the same dense-teacher and geometry-aware rewards.
    Lower is better for 3DCS and CamMC, while higher is better for IQ.
    }
    \label{tab:curriculum_ablation}
    \setlength{\tabcolsep}{5pt}
    \renewcommand{\arraystretch}{1.05}
    \resizebox{\linewidth}{!}{
    \begin{tabular}{lccc|ccc|ccc}
      \toprule
      & \multicolumn{3}{c|}{$s=0.125$}
      & \multicolumn{3}{c|}{$s=0.5$}
      & \multicolumn{3}{c}{$s=1$} \\
      Strategy
      & 3DCS $\downarrow$ & CamMC $\downarrow$ & IQ $\uparrow$
      & 3DCS $\downarrow$ & CamMC $\downarrow$ & IQ $\uparrow$
      & 3DCS $\downarrow$ & CamMC $\downarrow$ & IQ $\uparrow$ \\
      \midrule
      Direct large-step training
      & 0.6201 & 0.5104 & 0.4301
      & 0.6084 & 0.6287 & 0.4052
      & 0.5907 & 0.6621 & 0.4001 \\

      Non-cumulative curriculum
      & 0.6107 & 0.5028 & 0.4335
      & 0.6031 & 0.6314 & 0.4059
      & 0.5914 & 0.6643 & 0.4028 \\

      Mixed training
      & 0.6008 & 0.4976 & 0.4312
      & 0.5924 & 0.6263 & 0.4098
      & 0.5869 & 0.6658 & 0.4046 \\

      Cumulative curriculum (Ours)
      & \textbf{0.5868} & \textbf{0.4883} & \textbf{0.4398}
      & \textbf{0.5866} & \textbf{0.6154} & \textbf{0.4137}
      & \textbf{0.5825} & \textbf{0.6597} & \textbf{0.4071} \\
      \bottomrule
    \end{tabular}
    }
  \end{table}

\subsection{More Ablation Studies}
\textbf{Dense Trajectory Reward Ablation.} Table~\ref{tab:dense_teacher_ablation} evaluates the contribution of the two
  components in the dense-teacher reward. Using either depth or feature
  supervision improves 3DCS and CamMC over the base model, confirming that
  pose-aligned teacher observations provide useful supervision for sparse
  trajectories. The depth-based term yields a larger improvement in 3DCS,
  indicating that it better captures scene-layout and geometric discrepancies,
  whereas the feature-based term provides a stronger gain in imaging quality
  under larger motion. Combining both terms achieves the best overall geometric
  and camera-following performance, showing that depth and feature supervision
  provide complementary information.

\textbf{Geometry Reward Ablation.}
  Table~\ref{tab:geometry_reward_ablation} studies the contribution of the two
  geometry-aware rewards. Adding the reprojection reward mainly improves 3DCS,
  demonstrating its effectiveness in suppressing cross-frame structural drift
  and inconsistent parallax. In contrast, the relative-trajectory reward
  produces a larger gain in CamMC, indicating that it provides more direct
  supervision for camera-motion following. Combining both rewards achieves the
  best overall performance across motion magnitudes, confirming that
  cross-frame reprojection and trajectory-level evaluation provide complementary
  geometric signals.

\textbf{Curriculum Learning Ablation.}
 Table~\ref{tab:curriculum_ablation} compares different strategies for
  organizing motion difficulty during training. Directly training on the
  largest motion interval performs poorly on dense trajectories, indicating
  that aggressive motion expansion can cause capability forgetting. The
  non-cumulative curriculum improves optimization stability but still loses
  some performance because previously introduced easier trajectories are
  discarded at later stages. Mixing all motion intervals from the beginning
  provides a stronger balance, but is less effective than progressive training.
  The cumulative curriculum achieves the best overall results across all
  motion magnitudes, showing that retaining easier trajectories while
  gradually introducing larger motions is important for both motion-range
  expansion and capability preservation.

\end{document}